\documentclass[conference,letterpaper]{IEEEtran}
\IEEEoverridecommandlockouts
\usepackage{cite}
\usepackage{amsmath,amssymb,amsfonts}
\usepackage{algorithmic}
\usepackage{graphicx}
\usepackage{textcomp}
\usepackage{multirow}
\usepackage{booktabs}
\usepackage{subcaption}
\usepackage{microtype}
\usepackage{hyperref}
\usepackage{cleveref}
\usepackage{float}

\usepackage{comment}

\newcommand{\Stratega}{\textsc{Stratega}~}

\usepackage{xcolor}
\def\BibTeX{{\rm B\kern-.05em{\sc i\kern-.025em b}\kern-.08em
    T\kern-.1667em\lower.7ex\hbox{E}\kern-.125emX}}
    
\begin{document}

\title{Portfolio Search and Optimization for \\General Strategy Game-Playing}

\author{
\IEEEauthorblockN{Alexander Dockhorn}
\IEEEauthorblockA{\textit{School of Electronic Engineering and Computer Science} \\
\textit{Queen Mary University of London, UK}\\
London, UK \\
a.dockhorn@qmul.ac.uk}
\and
\IEEEauthorblockN{Jorge Hurtado-Grueso}
\IEEEauthorblockA{\textit{School of Electronic Engineering and Computer Science} \\
\textit{Queen Mary University of London, UK}\\
London, UK \\
a.dockhorn@qmul.ac.uk}
\and
\IEEEauthorblockN{Dominik Jeurissen}
\IEEEauthorblockA{\textit{School of Electronic Engineering and Computer Science} \\
\textit{Queen Mary University of London, UK}\\
London, UK \\
a.dockhorn@qmul.ac.uk}
\and
\IEEEauthorblockN{Linjie Xu}
\IEEEauthorblockA{\textit{School of Electronic Engineering and Computer Science} \\
\textit{Queen Mary University of London, UK}\\
London, UK \\
a.dockhorn@qmul.ac.uk}
\and
\IEEEauthorblockN{Diego Perez-Liebana}
\IEEEauthorblockA{\textit{School of Electronic Engineering and Computer Science} \\
\textit{Queen Mary University of London, UK}\\
London, UK \\
a.dockhorn@qmul.ac.uk}
}

\author{
\IEEEauthorblockN{Alexander Dockhorn, Jorge Hurtado-Grueso, Dominik Jeurissen, Linjie Xu, Diego Perez-Liebana}
\IEEEauthorblockA{\textit{School of Electronic Engineering and Computer Science} \\
\textit{Queen Mary University of London, London, UK}\\
\{a.dockhorn, diego.perez\}@qmul.ac.uk}
}

\IEEEoverridecommandlockouts
\IEEEpubid{\makebox[\columnwidth]{\hfill} 
\hspace{\columnsep}\makebox[\columnwidth]{ }}

\maketitle

\IEEEpubidadjcol

\begin{abstract}
    Portfolio methods represent a simple but efficient type of action abstraction which has shown to improve the performance of search-based agents in a range of strategy games.
    We first review existing portfolio techniques and propose a new algorithm for optimization and action-selection based on the Rolling Horizon Evolutionary Algorithm.
    Moreover, a series of variants are developed to solve problems in different aspects.
    We further analyze the performance of discussed agents in a general strategy game-playing task.
    For this purpose, we run experiments on three different game-modes of the \Stratega framework.
    For the optimization of the agents' parameters and portfolio sets we study the use of the N-tuple Bandit Evolutionary Algorithm.
    The resulting portfolio sets suggest a high diversity in play-styles while being able to consistently beat the sample agents.
    An analysis of the agents' performance shows that the proposed algorithm generalizes well to all game-modes and is able to outperform other portfolio methods.
\end{abstract}


\section{Introduction}

Digital real-time strategy games (RTS) represent a challenging genre for the development of artificial intelligence (AI). 
In RTS games, an AI agent is usually tasked to control a large set of units along a battlefield over a prolonged time-span. 
The difficulty arises from the large set of actions available to each unit and the complexity of controlling all of the units at the same time.
As a result, the game tree complexity of a game like Starcraft significantly exceeds the complexity of Go.

A possible method for addressing the complexity of a strategy game's action space is the use of action abstractions.
The most basic abstraction is a \textit{script}, which given a game-state and a unit, selects an appropriate action to be executed.
Developing a well-performing script is a demanding task in terms of time and complexity.
Instead of implementing a single, complex and hand-written rule-based player, search-based agents can make use of a \textit{portfolio}, formed by a set of simple scripts, to guide their search.
Meaning, the search-based agent chooses the next script which selects a suitable action instead of searching through the original action space.

The use of a portfolio fixes the number of available options per turn (one action per script), and therefore, limits the breadth of the search tree.
As a result, we can increase the search-depth when given the same amount of computation time.
Portfolio methods have been tested in games like Starcraft \cite{Churchill2013} and MicroRTS \cite{Moraes2018NestedGreedySF}, and have shown to significantly improve a search-based agent's performance.

With the recent advancements in strategy game AI, e.g. the success of AlphaStar\cite{Vinyals2019}, we believe that the time is ripe to approach general strategy game-playing.
In contrast to single game-playing, general game-playing requires an agent to perform well in several games.
Search-based approaches have shown to perform well in general game-playing tasks, such as the games of the general video game AI (GVGAI) competition~\cite{perez2019general} or Stanford's general game-playing competition~\cite{genesereth2005general}.
In this work, we want to test if portfolio methods are up to the task of playing a set of strategy games with common characteristics. 
For this purpose, we developed \Stratega \cite{Dockhorn2020, Stratega}, a general strategy games framework, which allows the definition of a wide variety of strategy games using a common API.

In this study, we want to propose and compare portfolio-based search algorithms for playing general strategy games.
Our contributions can be summarized by:
\begin{itemize}
    \item \textbf{Portfolio RHEA:} We combine portfolio online evolution with the rolling horizon evolutionary algorithm to propose a new kind of portfolio algorithm. Furthermore, we propose a multi-objective variant and a sparse encoding of genomes for long-time consistency of script assignments.
    \item \textbf{Optimization of portfolios and parameter sets:} We use the N-tuple bandit evolutionary algorithm to tune the parameter sets and portfolio composition for each game-mode. Our evaluation shows that the optimization procedure is able to detect varying successful play-styles.
    \item \textbf{Comparing performance and play-styles:} We compare the performance of reviewed and proposed portfolio methods in a separate round-robin tournament. Additionally, we analyze the agents' portfolio usage profiles with respect to the three game-modes to study their varying play-styles.
\end{itemize}


In \Cref{sec:Stratega} we review the \Stratega framework and its respective game-modes.
An introduction to portfolio-based search methods and the scripts used for playing \Stratega games is presented in \Cref{sec:portfolio}.
In \Cref{sec:prhea} we propose new portfolio methods based on the rolling horizon evolutionary algorithm. 
Our proposed optimization procedure will be covered in \Cref{sec:optimization}.
The agents' performance and play-styles are evaluated in three game-modes in \Cref{sec:performance}.
We conclude the paper with an outlook of future work in \Cref{sec:conclusion}.

\begin{figure*}
     \begin{subfigure}[b]{0.49\textwidth}
         \centering
         \includegraphics[width=\textwidth, clip, trim=0 1cm 0 0.5cm]{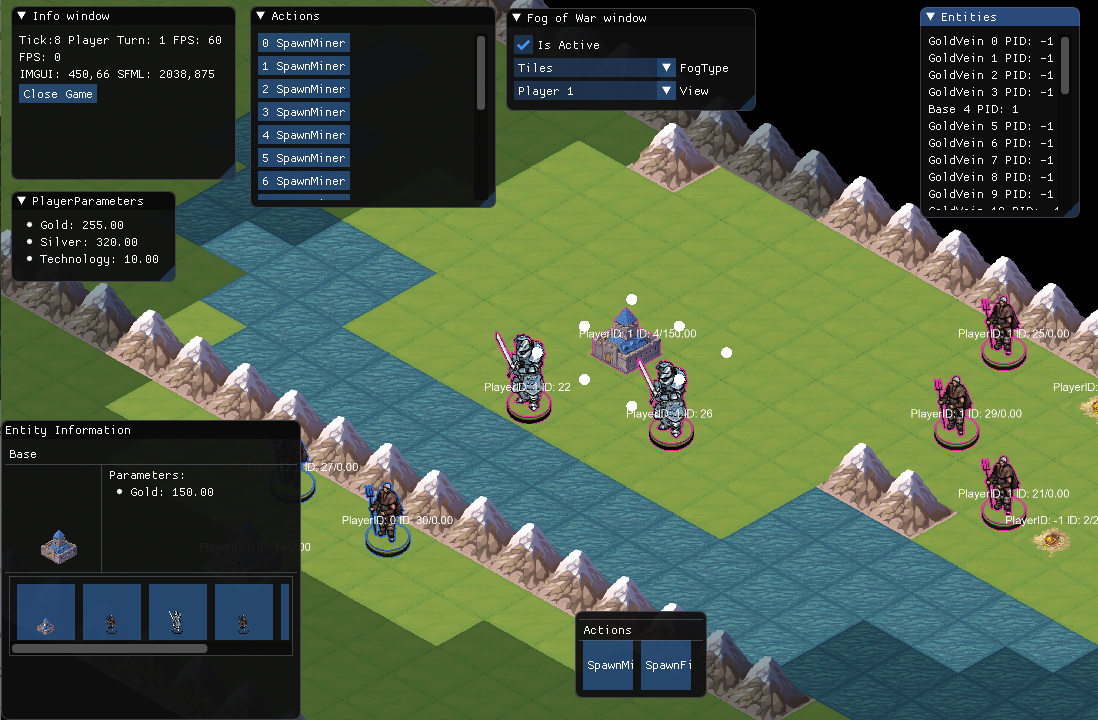}  
         \caption{A screenshot of the framework's current GUI.}
         \label{fig:screenshot}
     \end{subfigure}
     \hfill
     \begin{subfigure}[b]{0.49\textwidth}
         \centering
         \includegraphics[width=\textwidth, clip, trim=1cm -1em 0cm 0]{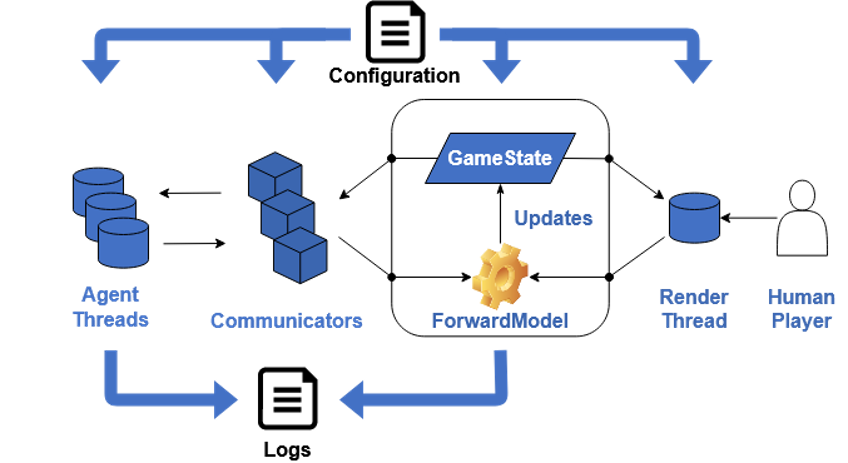} 
         \caption{Structure of the framework.}
         \label{fig:stratega-framework}
     \end{subfigure}
     \vspace{1em}
    \caption{Overview of the \Stratega framework.}
    \label{fig:stratega}
\end{figure*}

\section{Stratega}
\label{sec:Stratega}

\Stratega is a framework for studying AI development for general strategy games~\cite{Stratega}.
It allows the implementation of turn-based and real-time strategy games similar to games like Fire Emblem\footnote{Nintendo, 1990} or Starcraft.
Games are represented from a top-down view showing the battlefield.
Each player controls a set of units, which can each be moved along the battlefield, attack other units, and make use of their special abilities.

The \Stratega framework's components and its GUI are shown in \Cref{fig:stratega}. 
Using the YAML markup language, users can configure a game's components, including but not limited to, map tiles, units, and their abilities as well as the agents and their parameters.
Game communicators allow the agents to observe the current game-state.
In future iterations of the framework, they will be used as access point for developing agents in other programming languages.
The framework's graphical user interface allows the developer to play the game and debug developed agents by showing information on the game-state and agents in several floating windows. 
Additionally, the framework actively supports logging of game-play statistics and profiling of developed agents and game-modes.

In this work, we focus on turn-based, grid-based, multi-unit, multi-action games with two players.
Such a game can be described as a zero-sum two-player game in which the current player can execute an action to move the game forward.
Each unit has a range of actions it can execute, some popular examples include moving, attacking, and using abilities like healing another unit.
Executing any of these actions, will change the game-state and reduce the number of the units remaining actions for this turn by one.
Since the game allows a player to control multiple units, players can choose the unit and its action that should be executed next. 
At each given point in time, the current player can choose to end their turn, which also ends automatically in case none of the player's units can execute any actions.

\Stratega comes with a variety of pre-implemented game-modes.
In this work, we will be testing the implemented agents in three turn-based game-modes, each highlighting a different aspect of the framework, and therefore, posing a different challenge to the tested agents. 
\begin{itemize}
    \item The game-mode \textit{Kings} shares similarities with chess. Here, the players control a set of units and try to defeat the opponent's king. Each unit is unique in its attributes, including health, attack damage, movement and attack range. When a player's king is killed, the player immediately loses the game.
    \item Our second game-mode \textit{Pushers} requires the agents to use the unit's special ability for pushing another unit into nearby holes. Agents will need to avoid holes to stay alive. Using multiple pushes in the same turn can allow killing other units quickly.
    \item The game-mode \textit{Healers} shifts the focus to the healer unit and its ability to replenish its own and another unit's health. At the end of each turn, all units receive a fixed amount of damage. This forces the agents to quickly group their units and make effective use of the healer's ability. The player whose units survive the longest wins the game.
\end{itemize}

All games in \Stratega are played through a common interface for AI agent development, which provides access to the game's forward model (FM). An FM allows agents to simulate the outcome of their actions by providing a potential future state after supplying a state-action pair.
\textsc{Stratega}'s components have been optimized for speed, to allow as many simulations as possible per second.
Using a \textit{Windows 10 x64, with CPU: I7-6700HQ 2.60GHz; RAM: 16Gb; GPU GTX960m} the simulation allows for $\approx 100\,000$ forward model calls per second in turn-based mode and  $\approx 35\,000$ calls per second in real-time strategy mode.

More information on our project can be found under:
\begin{center}
\vspace{-0.5em}
\url{https://gaigresearch.github.io/afm/}
\end{center}
\vspace{-0.5em}

The current state of the framework can be accessed at:
\vspace{-0.5em}
\begin{center}
\url{https://github.com/GAIGResearch/Stratega}
\end{center}

\newpage
\section{Portfolio-based Search Methods}
\label{sec:portfolio}

\subsection{Portfolio-based Action Abstractions}

A \textit{script} is a function mapping a state and a unit to an action. 
Portfolio-based methods make use of a set of scripts to determine the suitable actions among all the actions available to the agent.
While traditional search algorithms navigate through the action space, portfolio methods search through the script space.
Therefore, each decision is reduced to the script that should be applied next.

For our evaluation, we implemented six simple scripts, which focus on different aspects of the game. 
\begin{itemize}
    \item \textbf{Attack Closest}: Attack the closest opponent unit in range. If no unit is close, walk to the closest opponent. In case no enemy is visible, act randomly.
    \item \textbf{Attack Weakest}: Attack the weakest opponent unit in range. If no unit is close, walk to the weakest known opponent unit. In case no enemy is visible, act randomly.
    \item \textbf{Run Away}: Walk to the tile that maximizes the distance to all known opponent units. In case no enemy is visible, act randomly.
    \item \textbf{Run To Friends}: Walk to the tile that minimizes the distance to all friendly units. In case no friendly unit exists, act randomly.
    \item \textbf{Use Special Ability} Use any special ability of this unit. Choose a random action in case no special ability is available.
    \item \textbf{Random}: Use a random action. Required to ensure a small chance of selecting any action, and therefore, ensures that the whole search space can be explored.
\end{itemize}

\subsection{Portfolio Greedy Search (PGS)} 

PGS is an any-time greedy search algorithm that assigns a script to each of the player and the opponent's units. 
It has been initially designed for RTS combat micro~\cite{Churchill2013} to tackle efficient decision making in complex search spaces. 
Instead of maximizing the number of states explored during a search, PGS narrows down the search to actions returned by scripts. 
Using a hill-climbing procedure, the player first optimizes the initial script assignment of the player's units. 
Afterwards, the opponent's script assignment is improved to model a stronger opponent. The process is repeated multiple times to improve the overall quality of the agent's response. PGS has shown to perform well in several strategy games\cite{Churchill2013,Moraes2018NestedGreedySF} and represents one of the baseline algorithms for our evaluation.

In previous work, it has been shown that the alternating optimization scheme of PGS may not converge.
This can occur because of the agent's improve procedure of the opponent's and the player's script assignments may be trapped in a never-ending cycle of optimizations (cf. \cite{Moraes2018NestedGreedySF}).
This can be avoided by only optimizing the player's actions. 
However, this requires knowledge about the opponent's strategy to be accurate.
If this is not the case, the player's improvement procedure is highly exploitable.

\subsection{Portfolio Online Evolution (POE)} 

POE replaces the hill-climbing procedure of PGS with evolutionary optimization. 
Here, each individual encodes the script assignment for units in the current game-state.
An individual's fitness is determined by simulating a play-out of fixed length and applying a heuristic in the case of non-terminal game-states. 
During a simulation, the opponent's actions will be selected by sampling from a fixed script.

At the start of each game turn, the agent generates a population of random individuals (unit-script assignments).
Those are iteratively improved by mutating and recombining the fittest individuals. 
For mutating individuals, we chose a uniform mutation that randomly replaces a unit's script assignment with probability $p$.
The crossover operator recombines the script assignments of two individuals, by randomly choosing a parent from which it samples the script assignment.
Every time a unit's action has been executed, the current population will be reused until the end of the turn.

Similarly to PGS, POE has shown to perform well in several strategy games\cite{justesen2016online,justesen2017playing}.
They both represent the baseline algorithms in our evaluation of portfolio optimizing agents.
A comparison of portfolio-based search methods has recently been published by Moraes et al.~\cite{Lelis2020}. 
It presents a unifying perspective of PGS and POE as instances of the General Combinatorial Search for Exponential Spaces (GEX) algorithm.

\section{Portfolio Rolling Horizon Evolutionary Algorithm (PRHEA)}
\label{sec:prhea}

While POE optimizes the unit-script assignment for a single turn, we want to propose a different optimization strategy based on the rolling horizon evolutionary algorithm (RHEA)~\cite{perez2013rheaptsp}.
Since many turn-based strategy games allow the agents to observe the result of a single unit's action, we can use the new observation to reevaluate the agent's unit-script assignment. 
This is especially important in case the game-state is only partially observable, such that fog of war covers all tiles that are not in the immediate field of vision of one of the agent's units. Moving a unit may uncover new opponents that require the agent to react accordingly.

In games featuring a single action per turn, individuals of the RHEA algorithm encode a sequence of unit actions with a fixed length. 
After the sequence has been optimized by an evolutionary algorithm, the first action of the best individual is executed.
Furthermore, the action is removed from the sequence and a random action is appended at the end of the sequence.
The resulting individual is used to initialize a new population of units, which will be used to search for the best action in the next turn or game frame.
RHEA has already shown good playing performance in the context of general game-playing~\cite{Gaina2017-RHEA,Dockhorn2020Diss,gvgaibook2019}.

In PRHEA for multi-action turn-based games, we replace the individuals encoding with a sequence of scripts instead of actions.
Hence, an individual can be evaluated by executing all the actions returned by the sequence of scripts,
whereas a script is applied to the unit with the lowest ID that has actions left.
In the following, we propose two specialized variants of PRHEA.

\begin{figure*}[thb]
    \centering
     \begin{subfigure}[b]{0.49\textwidth}
         \centering
         \includegraphics[height=1.85cm, clip, trim=0 0cm 0 0cm]{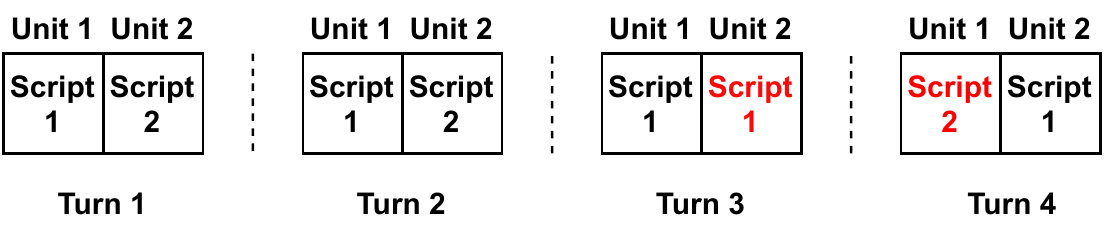}
         \caption{dense unit script assignment for 2 units and 4 turns}
         \label{fig:sphrea-dense}
     \end{subfigure}
    \hfill
      \begin{subfigure}[b]{0.49\textwidth}
         \centering
         \includegraphics[height=1.85cm, clip, trim=0 0cm 0 0cm]{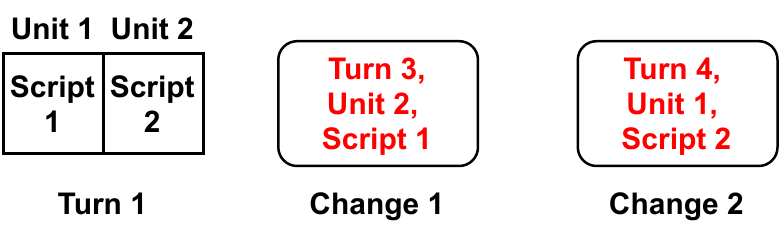}
         \caption{incremental script assignment and its changes over time}
         \label{fig:sphrea-sparse}
     \end{subfigure}
        \caption{Comparison of encoding strategies of unit script assignments.}
        \label{fig:encoding-comparison}
\end{figure*}

\subsection{Multi-objective Portfolio RHEA (MO-PRHEA)}

Strategy games may require an agent to chase multiple goals at once, e.g. killing the opponent's units while keeping your own units alive.
We want to reflect this multi-objective optimization problem, by proposing a multi-objective variant of the PRHEA algorithm, called MO-PHREA.
To test the approach, we implemented the non-dominated sorting genetic algorithm (NSGA-2)~\cite{nsga2} for optimizing the script assignment sequences of PRHEA.

All other agents rely on a single heuristic that considers the strength and the health of the player's and the opponent's units.
Due to the limited search depth of these agents, we have observed that fights on large maps often result in a draw since the agents are not able to find each other.
We have chosen to implement a second heuristic, that measures the average distance of the player's units to the opponent's units.
Whereas the first heuristic should be maximized, the agent will try to minimize the average distance measured by the second heuristic.
During the optimization, NSGA-2 will try to find a Pareto-front for both heuristics.
Nevertheless, when the agent selects the final action, it will prefer individuals that score well in the first heuristic. While this has shown to be the best performing strategy for the tested games, other selection strategies may allow to implement different agent personas.

\subsection{Sparse Portfolio RHEA (S-PRHEA)}

In a preliminary evaluation of agents introduced above, we observed that POE and PRHEA struggle with optimizing long action sequences.
A common scheme we observed is that the agents did not consistently use their units. 
While strong attackers had been used to attack the opponent and run back to defend the player's units, healers had been used to heal and attack.
A higher efficiency would have been achieved if the agent had stuck with the same script assignment for a prolonged time-frame.

To plan long sequences, we propose the S-PRHEA algorithm which uses two separate genomes for a single individual.
The first genome encodes a unit-script assignment similar to POE.
In its second genome, S-PRHEA encodes script assignment changes. 
Each change includes three variables, (1) the ticks left until this script change should be executed, (2) the unit that will be affected, and (3) the new script assigned to the unit.
Every time an action is being executed the tick counter will be reduced by one.
In case it reaches 0, the script change will be applied to the first genome and a new random script change will be added to the list (cf. \Cref{fig:encoding-comparison}).

By keeping the script assignment constant in the absence of change events, the agent is now able to consistently select and execute a script for each unit.
In exchange, the search space is expanded by the possible combinations of the second genome.
Depending on the number of turns simulated, and the number of available units and scripts, it can become a daunting task to find a useful script change with the correct timing.
Therefore, it may be practical to dynamically balance the time spent on optimizing both genomes.
Another solution may be to use cooperative co-evolution for optimizing two separate populations, each containing genomes of the same type.

\section{Parameter and Portfolio Optimization}
\label{sec:optimization}

The methods reviewed and proposed in the previous sections have plenty of parameters that can be adjusted. Not just can the search behavior be modified through the algorithm's internal parameters, but we can also define the search space in terms of the set of scripts that are included in the agent's portfolio.
To approach this demanding task, we propose to use the N-Tuple Bandit Evolutionary Algorithm (NTBEA)~\cite{lucas2018n}.

\begin{figure*}[p]
    \centering
    \vspace{1em}
     \begin{subfigure}[b]{\textwidth}
         \centering
         \includegraphics[width=0.56\textwidth, clip, trim=0 1cm 5cm 0.5cm]{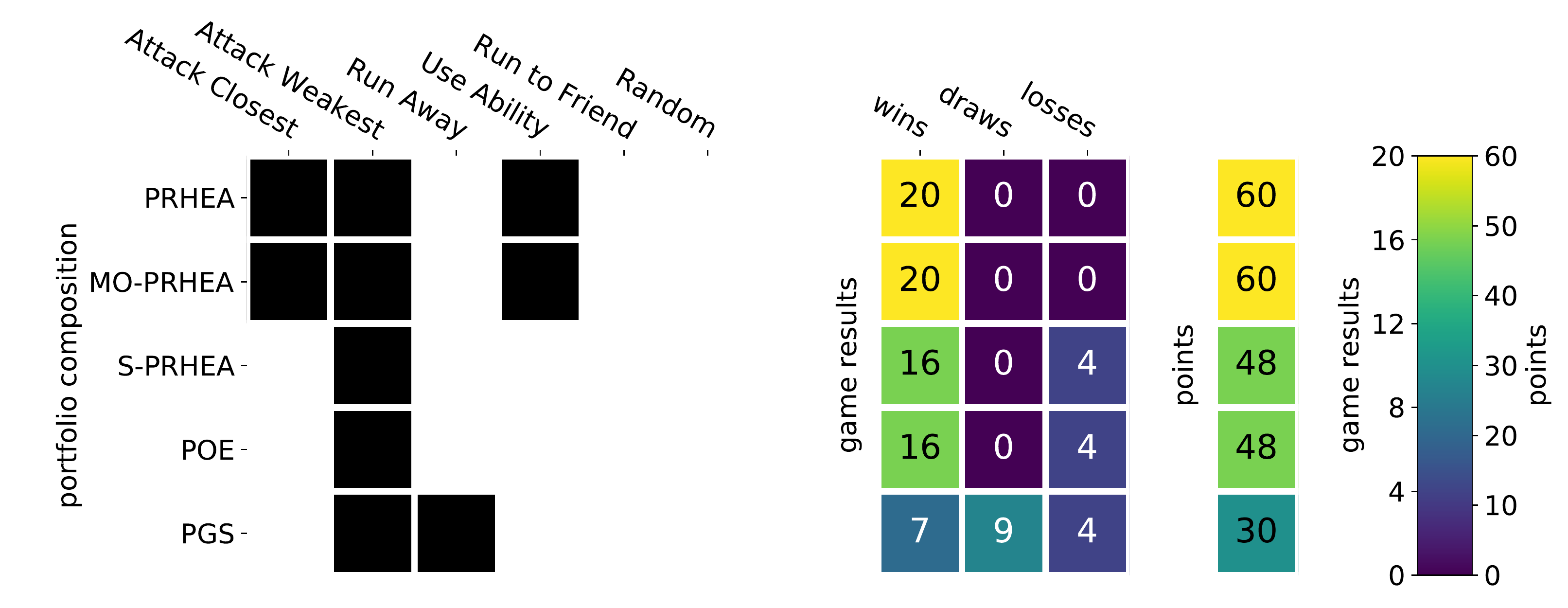}
         \hspace{1cm}
         \includegraphics[width=0.094\textwidth, clip, trim=28cm 1cm 0 0.5cm]{figures/Kings_updated.pdf}
         \caption{Kings}
         \label{fig:portfolio-kings}
     \end{subfigure}
     \vspace{0.5em}

      \begin{subfigure}[b]{0.49\textwidth}
         \centering
         \includegraphics[width=\textwidth, clip, trim=0 1cm 0 0.5cm]{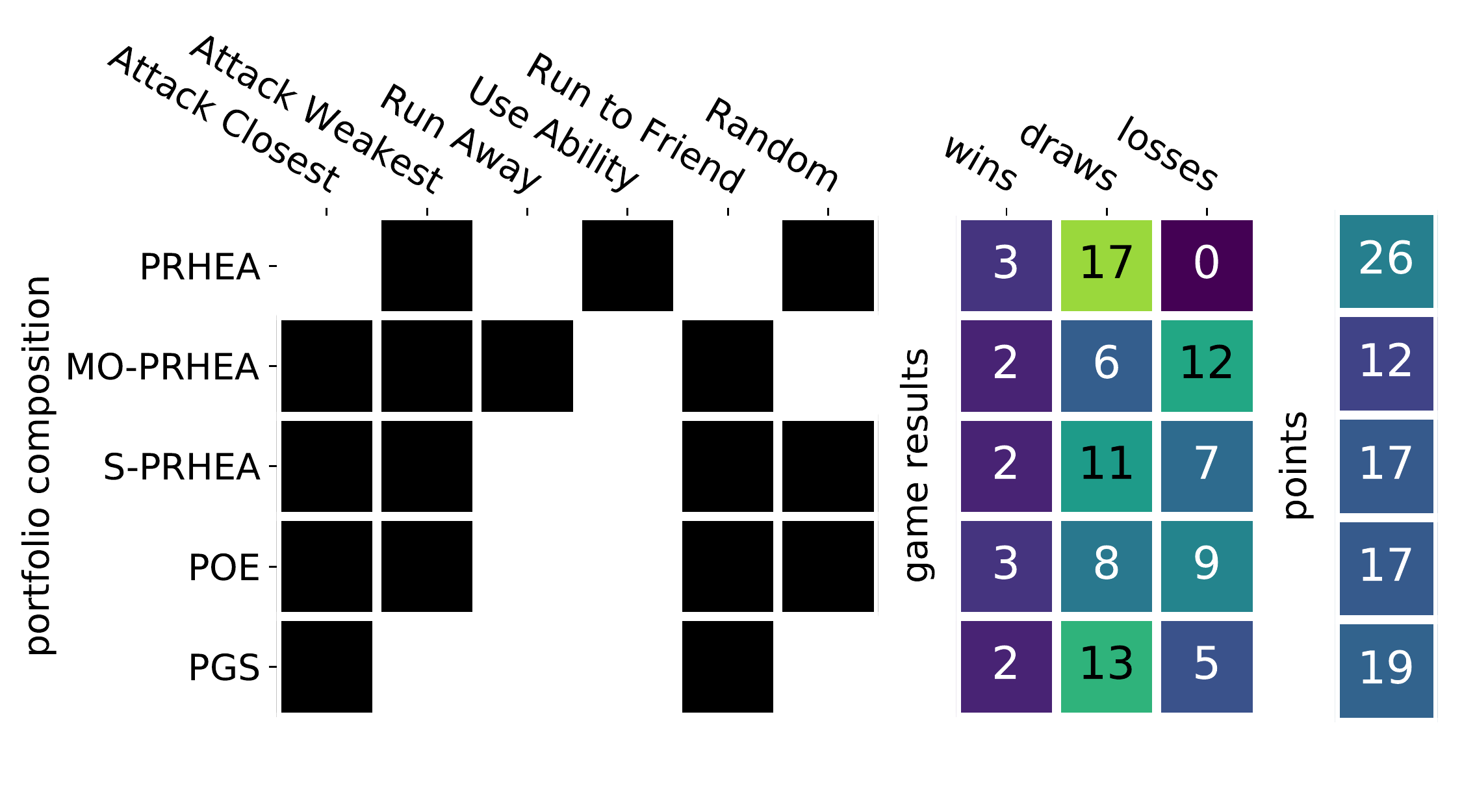}
         \caption{Pushers}
         \label{fig:portfolio-pushers}
     \end{subfigure}
     \hfill
     \begin{subfigure}[b]{0.49\textwidth}
         \centering
         \includegraphics[width=\textwidth, clip, trim=0 1cm 0 0.5cm]{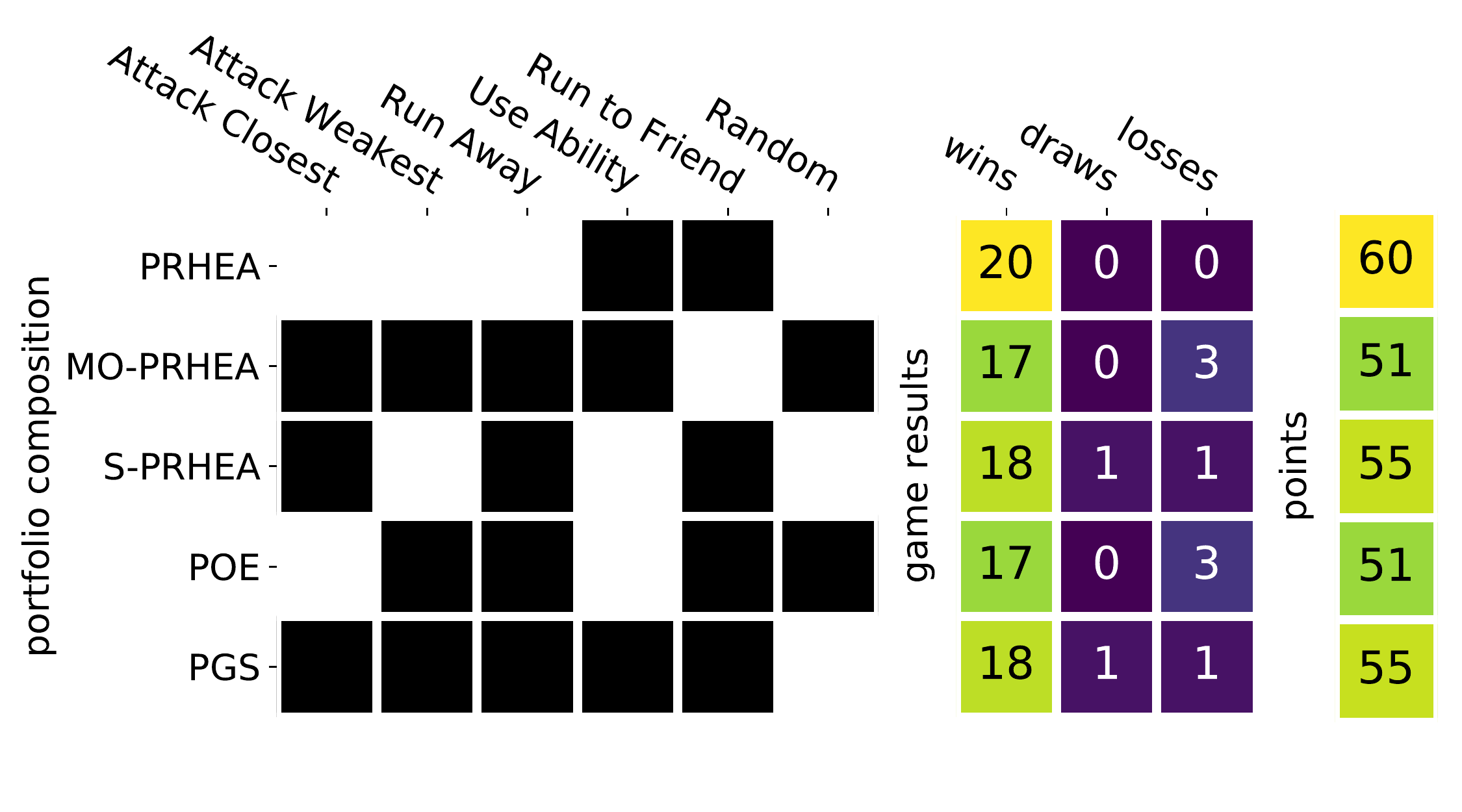}
         \caption{Healers}
         \label{fig:portfolio-healers}
     \end{subfigure}

        \caption{Overview of the algorithms' optimized portfolio sets and their performance against the rule-based agents throughout the three tested game-modes. A black square indicates that the script has been included.}
        \label{fig:portfolio-comparison}
\end{figure*}

For each pair of algorithm and game-mode, NTBEA was given the budget to evaluate $100$ parameter combinations.
The fitness of a selected parameter set is evaluated by simulating $20$ games against a rule-based agent.
The framework's combat agent has been used as an opponent for the game-modes \textit{Healers} and \textit{Kings}.
It focuses on attacking isolated units while healing the agent's own units.
When attacking or healing a unit, the agent prioritizes units with high strength.
The rule-based pushers agent has been used for its respective game-mode and uses path-planning to find ways for pushing opponent units into a trap.
A detailed description of the rule-based agents can be found in our previous paper~\cite{Stratega}.
The fitness is increased by $3$ for each win and remains unchanged after a loss.
After $100$ turns the game is automatically terminated and it results in a draw.
In this case, the agent's fitness is increased by $1$.
The turn limit has proven to be mostly sufficient since the mean number of turns until a win or loss has been observed to be $29.07$ (Kings), $6.27$ (Healers), and $3.48$ (Pushers) respectively.
The result of our optimization procedure is show in Tables \ref{tab:parameters-kings}-\ref{tab:parameters-healers}.
We separately compare the portfolios in \Cref{fig:portfolio-comparison}.

\begin{table*}[p]
    \centering
    \caption{Results of the NTBEA parameter optimization for the game mode \textit{Kings}}
    \label{tab:parameters-kings}
    \begin{tabular}{c|cccccccc}
        \toprule
        \multirow{2}{*}{Kings} & Population & Individual & Mutation & Tournament & Number of  &  Response & \multirow{2}{*}{Elitism} & Continue \\
        & Size      & Length & Rate & Size & Changes & Iterations & & Search \\
        & $\lbrace 1, 10, 100\rbrace$ & $\lbrack 1, 10\rbrack$ & $\lbrace 0.1, 0.5, 0.9\rbrace$ & $\lbrace 3, 5, 10 \rbrace$ & $\lbrack 1, 3, 5, 10\rbrack$ & $\lbrack 1, 5\rbrack$ & $\{\times, \checkmark\}$ & $\lbrace\times, \checkmark\rbrace$ \\
        \midrule
         PRHEA      &   1 &   1 & 0.5 &   5 & ---& ---& $\checkmark$ & $\checkmark$\\
         MO-PRHEA   &   1 &   1 & 0.1 &   5 & ---& ---& $\checkmark$ & $\checkmark$\\
         S-PRHEA    &   1 &   3 & 0.5 &   5 &  5 & ---& $\checkmark$ & $\checkmark$\\
         POE        &   1 &   3 & 0.5 &   5 & ---& ---& $\checkmark$ & $\checkmark$\\
         PGS        &   ---  &   3 &  ---   &  ---   & --- & 4 & --- & ---  \\
         \bottomrule
    \end{tabular}
    \vspace{1.5em}

    \centering
    \caption{Results of the NTBEA parameter optimization for the game mode \textit{Pushers}}
    \label{tab:parameters-pushers}
    \begin{tabular}{c|cccccccc}
        \toprule
        \multirow{2}{*}{Pushers} & Population & Individual & Mutation & Tournament & Number of  &  Response & \multirow{2}{*}{Elitism} & Continue \\
        & Size      & Length & Rate & Size & Changes & Iterations & & Search \\
        & $\lbrace 1, 10, 100\rbrace$ & $\lbrack 1, 10\rbrack$ & $\lbrace 0.1, 0.5, 0.9\rbrace$ & $\lbrace 3, 5, 10 \rbrace$ & $\lbrack 1, 3, 5, 10\rbrack$ & $\lbrack 1, 5\rbrack$ & $\{\times, \checkmark\}$ & $\lbrace\times, \checkmark\rbrace$ \\
        \midrule
         PRHEA      & 100 &   3 & 0.5 &   5 & --- & --- & $\checkmark$ & $\checkmark$\\
         MO-PRHEA   & 100 &   1 & 0.1 &   3 & --- & --- &  $\checkmark$ & $\checkmark$\\
         S-PRHEA    & 100 &   3 & 0.1 &   3 &  5  & --- &  $\checkmark$ & $\checkmark$\\
         POE        &  10 &   5 & 0.9 &   3 & --- & --- &  $\checkmark$ & $\checkmark$\\
         PGS        &   ---  &   2 &   ---  &   ---  & --- &3& --- & ---\\
         \bottomrule
    \end{tabular}

    \vspace{1.5em}
    \caption{Results of the NTBEA parameter optimization for the game mode \textit{Healers}}
    \label{tab:parameters-healers}
    \begin{tabular}{c|cccccccc}
        \toprule
        \multirow{2}{*}{Healers} & Population & Individual & Mutation & Tournament & Number of  &  Response & \multirow{2}{*}{Elitism} & Continue \\
        & Size      & Length & Rate & Size & Changes & Iterations & & Search \\
        & $\lbrace 1, 10, 100\rbrace$ & $\lbrack 1, 10\rbrack$ & $\lbrace 0.1, 0.5, 0.9\rbrace$ & $\lbrace 3, 5, 10 \rbrace$ & $\lbrack 1, 3, 5, 10\rbrack$ & $\lbrack 1, 5\rbrack$ & $\{\times, \checkmark\}$ & $\lbrace\times, \checkmark\rbrace$ \\
        \midrule
         PRHEA      & 100 & 10 & 0.1 &   3 & --- & --- & $\checkmark$ & $\checkmark$  \\
         MO-PRHEA   &  10 &  1 & 0.1 &   5 & --- & --- & $\checkmark$ & $\checkmark$  \\
         S-PRHEA    & 100 &  1 & 0.1 &  10 &  5  & --- & $\checkmark$ & $\checkmark$  \\
         POE        &  10 &  3 & 0.1 &   3 & --- & --- & $\checkmark$ & $\checkmark$  \\
         PGS        & ---    &  3 &  ---   &  ---   & ---&1&--- & ---\\
         \bottomrule
    \end{tabular}
\end{table*}

Most interestingly are the different portfolios found for each algorithm and game-mode.
For the game-mode Kings, most algorithms have excluded the \textit{Run Away}, \textit{Run to Friend}, and \textit{Random} scripts.
While \textit{Run to Friend} can be helpful for the healer unit, it cannot effectively be used for other unit types.
All but the PGS agent are able to beat the combat agent consistently.
In the game-mode \textit{Healers}, the portfolios are more diverse.
The best performing agent, PRHEA, focuses on the two scripts \textit{Use Ability} and \textit{Run To Friend}.
As explained before, those are especially useful for using the healer unit effectively.
Other algorithms focus on their fighting capabilities and include attack actions.
Both strategies seem to be able to consistently beat the combat agent.
For the final game-mode pushers, agents were less successful. 
This has been an expected result since none of the scripts has been specifically designed for the \textit{Push} action. 
In contrast to the special ability \textit{Heal}, the \textit{Push} ability is only useful in case preconditions are satisfied, i.e. the opponent is standing next to a hole.
Therefore, the success rate of the \textit{Use Ability} script is generally lower than in the other game-modes.
Additionally, it is worthwhile remembering that the attacking scripts effectively become equal to the random script because none of the agent's units will be able to attack.

Similar interesting observations can be made for the algorithm's parameter sets.
In the combat scenarios of Kings, all agents prefer a population size of $1$, effectively using a $1+1$ evaluation.
In this case, we can ignore the tournament size since the agent solely relies on mutation.
When we look at the parameter sets of the game-modes \textit{Healers} and \textit{Pushers}, we can see that all algorithms prefer larger population sizes.
This becomes especially relevant in the game-mode \textit{Pushers} because the introduction of many random individuals per generation raises the chances of sampling a push action.
In the case of the parameters \textit{elitism} and \textit{continue search}, it is of no wonder that the algorithms prefer to reuse the result of the previous iteration during and after action selection.
This increases the agent's consistency and ensures that once a good action sequence has been found it can be executed entirely.
The S-PRHEA agent seems to prefer to store $5$ changes in its second genome. This has been the highest value in the search space, which suggests that the algorithm prefers more variation over time.

\begin{figure}[!ht]
    \centering
    \vspace{1em}

     \begin{subfigure}[b]{0.49\textwidth}
         \centering
         \includegraphics[width=\textwidth, clip, trim= 0 0 0 1.7cm]{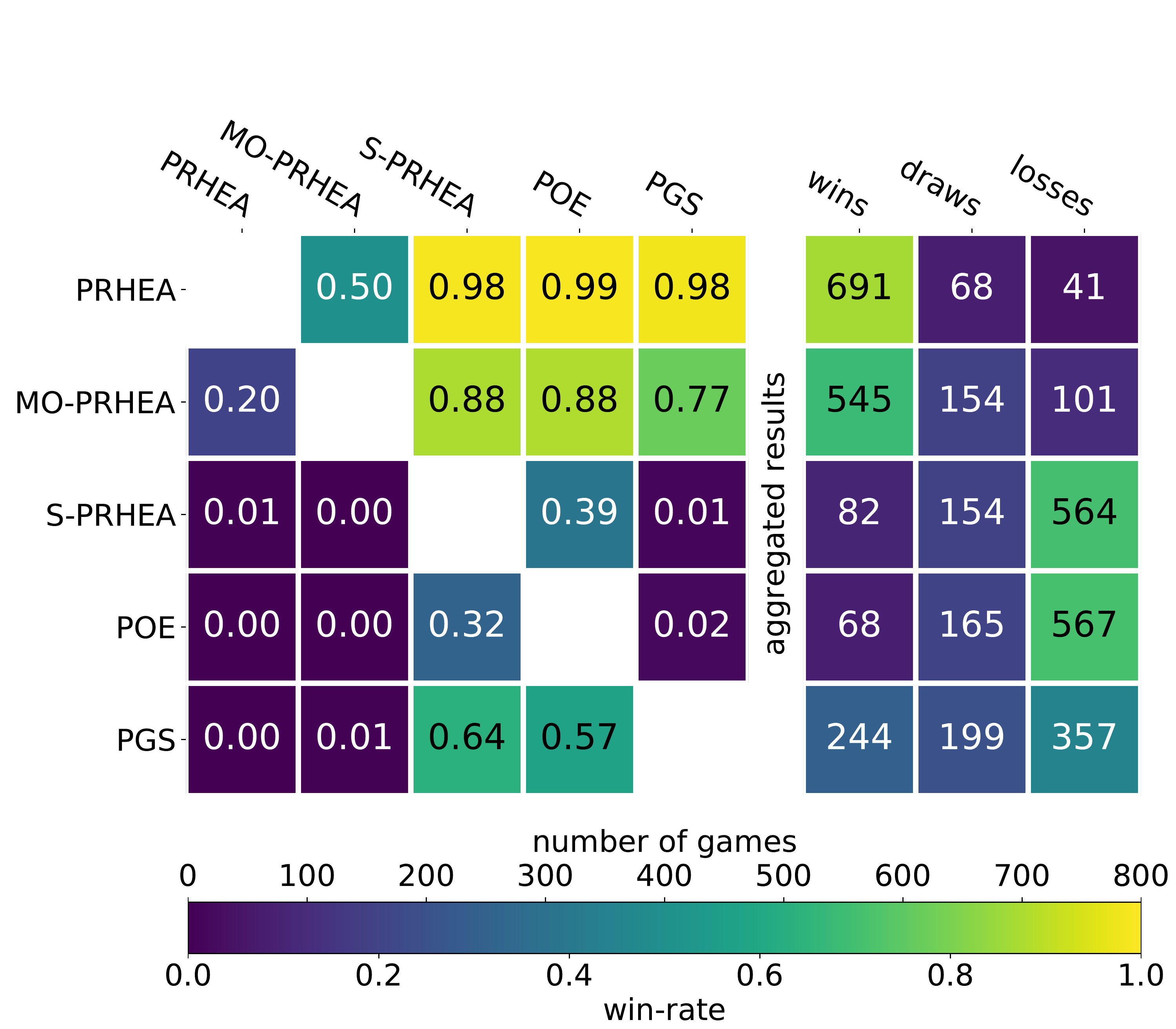}
         \caption{Kings}
         \label{fig:win-rate-kings}
     \end{subfigure}

    \begin{subfigure}[b]{0.49\textwidth}
         \centering
         \includegraphics[width=\textwidth, clip, trim= 0 6cm 0 1.8cm]{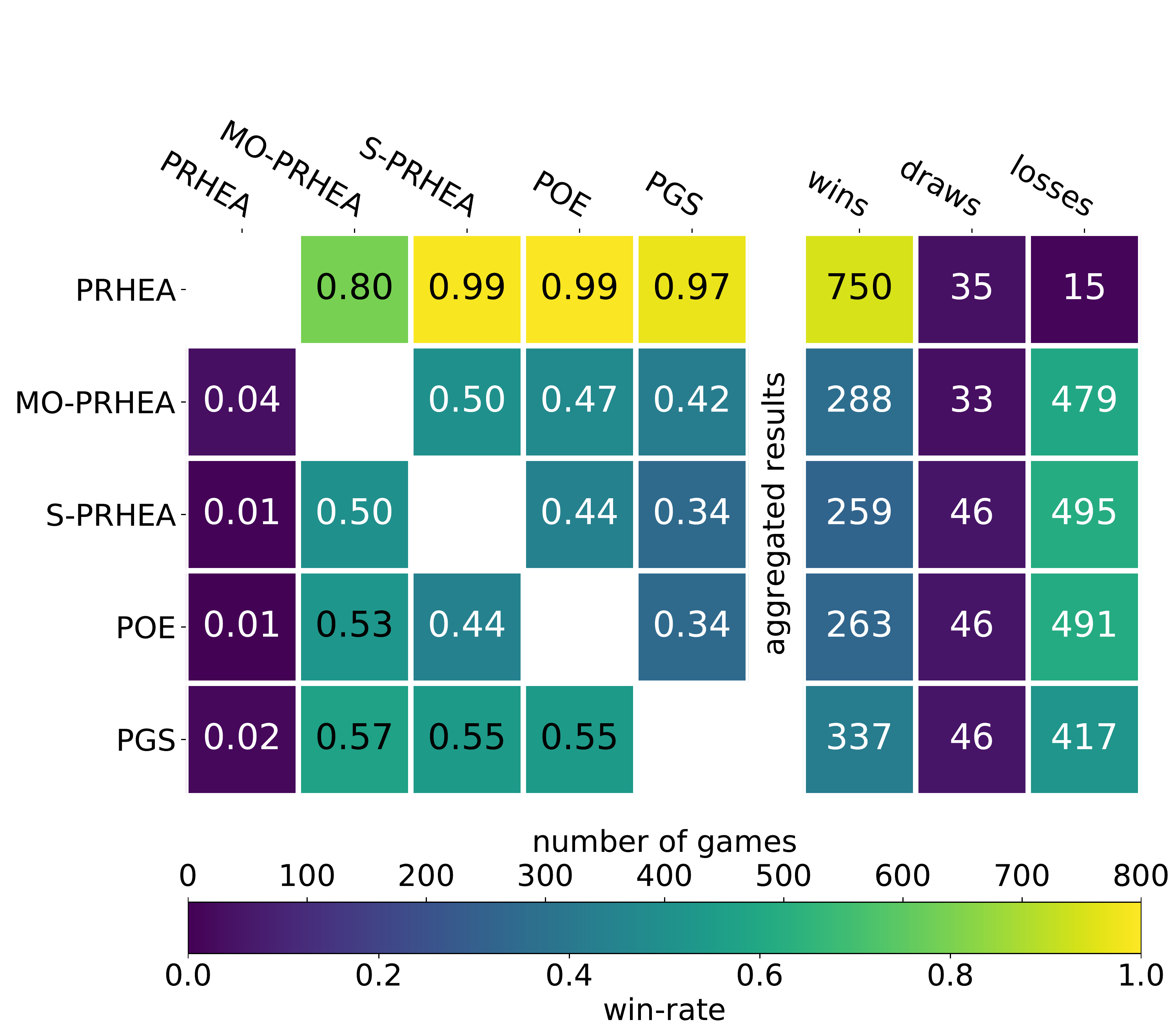}
         \caption{Pushers}
         \label{fig:win-rate-pushers}
     \end{subfigure}
     
     \begin{subfigure}[b]{0.49\textwidth}
         \centering
         \includegraphics[width=\textwidth, clip, trim= 0 6cm 0 1.9cm]{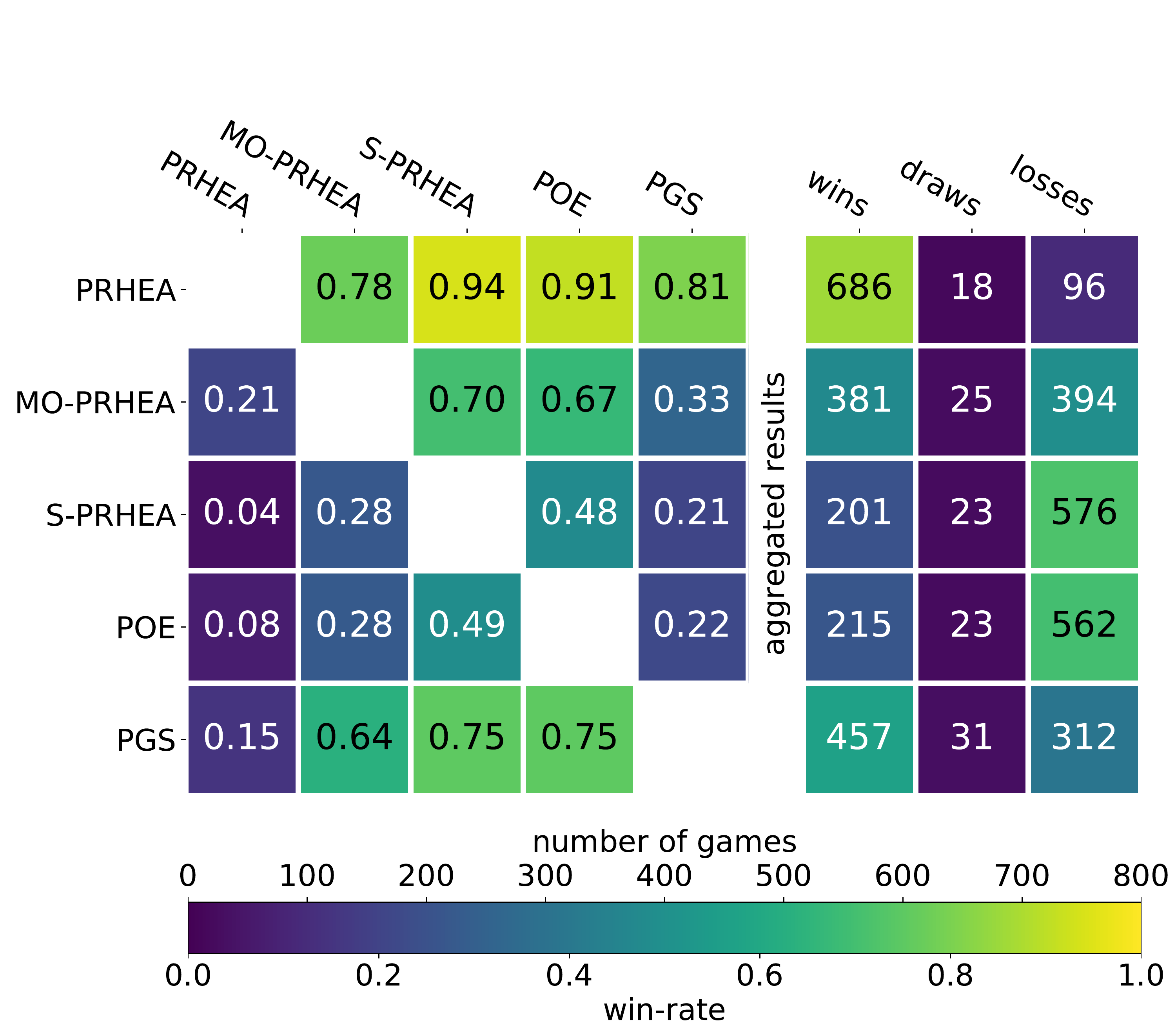}
         \caption{Healers}
         \label{fig:win-rate-healers}
     \end{subfigure}

    \caption{Performance comparison in a round-robin tournament of 200 games per match-up throughout the three tested game-modes. Each cell $(i,\,j)$ shows the win-rate of the agent in row $i$ against the agent in column $j$. The columns to the right show the total number of wins, draws, and losses per agent.}
    \label{fig:win-rate}
\end{figure}

\begin{figure*}[t!]
     \centering
     \begin{subfigure}{0.90\textwidth}
        \centering
         \includegraphics[width=\textwidth, clip, trim= 0 0cm 0 0.5cm]{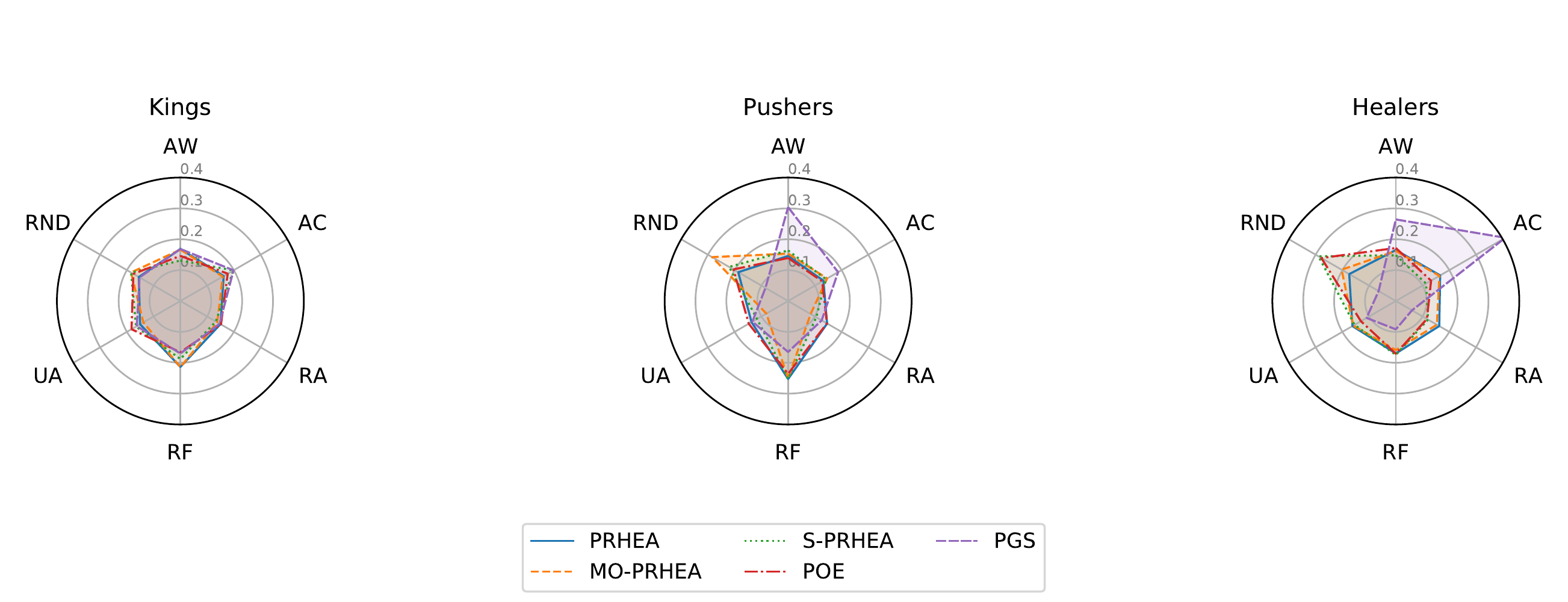}
         \caption{grouped by game-mode}
         \label{fig:portfolio-usage-gamemode}
     \end{subfigure}
    \vspace{1em}
    
     \begin{subfigure}{1.0\textwidth}
        \includegraphics[width=\textwidth, clip, trim= 0 0cm 0 0.5cm]{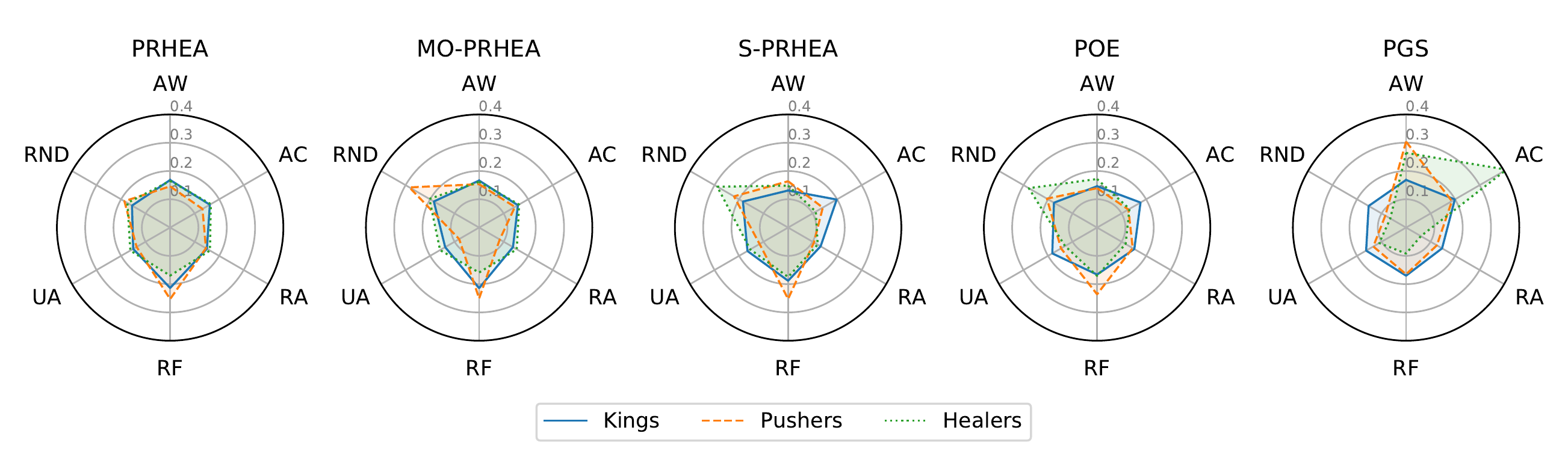}
        \caption{grouped by agent}
        \label{fig:portfolio-usage-agent}
     \end{subfigure}
     \caption{Portfolio usage profiles for each combination of agent and game-mode. Each axis shows the relative usage of a script while playing against the rule-based combat agent. Meaning of abbreviated axis descriptions in clockwise order starting from the top: \textit{AW} = attack weakest, \textit{AC} = attack closest, \textit{RA} = run away, \textit{RF} = run to friend, \textit{UA} = use ability, \textit{RND} = random script.}
     \label{portfolio-usage}
\end{figure*}

\section{Performance Comparison}
\label{sec:performance}

After having optimized the parameter sets, we compare the game-play performance of the introduced portfolio methods.
Therefore, we simulate a round-robin tournament of $200$ games per pair of agents and measure their win-rate.
Since the starting position is randomized and one player's starting position may be advantageous, we use the seeds 1 to 100 for simulating rematches with swapped starting positions.
Therefore, each map is played two times with interchanged roles.
The maximal length of a game has been set to $100$ turns.
In case no player has met the winning condition by then, the game results in a draw.
\Cref{fig:win-rate} shows a heatmap visualizing the win-rates of all tested agents and their total number of wins, draws, and losses. 
Each cell $(i,\,j)$ shows the win-rate of the agent in row $i$ against the agent in column $j$.

In all three game-modes, the PRHEA agent has scored the most wins and was able to win at least $50\%$ of its games against the other agents. 
This is in line with its results against the rule-based agents.
In the game-mode \textit{Healers}, we saw two strategies emerging against the rule-based agent.
While PRHEA has focused on healing and staying together as a group, other agents focused on winning by attacking the opponent.
Our evaluation shows that the former is the better strategy when matched against each other.
PRHEA wins $686$ out of $800$ games against the other agents, showing a clear superiority against every one of them.

The MO-PRHEA and the PGS agents have scored second best. 
In the game-mode \textit{Kings} MO-PRHEA has outperformed PGS, which is likely due to using the same portfolio as PRHEA.
In the other two game-modes, PGS performed surprisingly well considering that its score against the rule-based agent has not stood out.
This may be due to the opponent model used for the initialization of the unit-script assignment in PGS.
Here, we used the attack closest script, which often coincided with the made decisions by the opponent's agent.
Therefore, PGS has been able to outplay its opponents by choosing a valid counter-strategy.

Both S-PRHEA and POE did not perform well in comparison with other agents, despite their comparable performance against the rule-based agents.
This might be due to their portfolio or related to the algorithm itself.
We hope that we can gain more insights by further analysis of these two methods in future work.

\subsection{Portfolio Usage Profiles}

In a second evaluation, we want to get more insight into the agents' strategical decisions by comparing their portfolio usage when being provided with the same set of scripts.
Therefore, we simulated $1000$ games against the rule-based agents for each combination of portfolio agents and game-modes.
During the game, we record the usage statistics of a script each time it is used to return an action for execution.
\Cref{fig:portfolio-usage-gamemode} and \Cref{fig:portfolio-usage-agent} show a comparison of the agents' portfolio profiles.

It is interesting to see how the portfolio profiles vary per game-mode and agent.
In the game-mode \textit{Kings}, all agents tend to use the scripts uniformly.
This stands in stark contrast to the results we obtained by using NTBEA to optimize the portfolio sets, in which all agents have focused on the attack scripts.
In the case of the game-mode \textit{Healers}, we see \mbox{S-PRHEA} and POE exhibiting a profile that more often relies on the random action script.
This may be a weakness that also explains their relatively weak performance in our previous experiment.
Surprisingly, their profiles are very similar in all three game-modes.
In contrast, the PGS agent exhibits a profile that is very different from the other agents in the game-modes \textit{Healers} and \textit{Pushers}.
Here, the agent strongly favors the two attack-based scripts.
We expect that the difference of PGS stems from its initialization procedure in which the attack weakest script is assigned to all units.
Another obvious change in respect to the NTBEA optimized portfolio sets can be observed for the PRHEA agent. When being provided with all scripts, the PRHEA agent also uses a combat-oriented strategy instead of focusing on healing actions in the game-mode \textit{Healers}. 
For the game-mode \textit{Pushers}, we also see different portfolio profiles emerging. 
However, as mentioned before, multiple scripts will result in random actions being returned in case no applicable action can be found in the action space.
Therefore, observed differences are less meaningful than for other game-modes.

\section{Conclusion}
\label{sec:conclusion}

In this work, we have proposed three new portfolio methods based on the rolling horizon evolutionary algorithm, i.e. PRHEA, MO-PRHEA, and S-PRHEA. 
To optimize their parameter and portfolio sets we employed the NTBEA algorithm, which was able to uncover a variety of play-styles in the three tested game-modes of the \Stratega framework.
In a subsequent evaluation of the agents' performance, the PRHEA agent has shown to perform best.
Proposed variants have shown to perform well against the rule-based agent, but were not able to beat the baseline methods PGS and POE.
A comparison of the agents' portfolio usage profiles has shown that the agents favor different play-styles when being provided with the whole set of scripts.
This highlights the interesting opportunities of optimization methods like NTBEA for uncovering new play-styles by optimizing an agent's portfolio set.

In future work, we want to extend our analysis on portfolio optimization to enhance the variability of search-based agents. Algorithms like map-elites~\cite{DBLP:journals/corr/MouretC15} may help to find a diverse set of well-performing strategies. Additionally, we plan to analyze methods for script generation to explore the strategy space and complement an agent's portfolio.

In another line of work, we plan to analyze game-state abstractions to improve the search efficiency in large state spaces as they can be found in strategy games.
While action abstractions have been a common theme in recent research on portfolio methods, game-state abstractions, which have the potential to drastically reduce the search space, have not received the same attention.

Finally, as the \Stratega framework is still being developed, we will incorporate more complex games into this study, which will bring common features in strategy games such as economic management, technology trees, and build orders.

\newpage 
\section*{Acknowledgements}
This work is supported by UK EPSRC research grant EP/T008962/1 (\url{https://gaigresearch.github.io/afm/}).
\vspace{2em}

\bibliographystyle{bibliography/IEEEtran}
\bibliography{ref}

\vspace{12pt}

\end{document}